\documentclass[a4paper,english]{lipics-v2018}

\pdfoutput=1

\usepackage[final]{microtype}

\usepackage{xspace}
\usepackage{xparse}
\usepackage{etoolbox}

\usepackage{amsmath}
\usepackage{mathtools,thmtools}
\usepackage[capitalise]{cleveref}
\usepackage[all]{foreign}
\usepackage{array}
\usepackage{longtable}
\usepackage{mdwtab}
\usepackage{bigdelim}
\usepackage{wrapfig}
\usepackage{complexity}

\usepackage[only,llbracket,rrbracket]{stmaryrd}

\usepackage{tikz}

\usetikzlibrary{calc,decorations.pathreplacing,patterns}

\tikzset{
  math mode/.style={execute at begin node=$, execute at end node=$},
  every node/.style={font=\footnotesize}
}

\definecolor{lipics}{rgb}{0.99,0.78,0.07}

\usepackage{lipsum}

\newcommand*\termfont{\sffamily}
\DeclareTextFontCommand{\textterm}{\termfont}

\newcommand*\defterm[2]{%
  \DeclareDocumentCommand{#1}{s}{%
    \textterm{#2}\IfBooleanT{##1}{\textsuperscript{*}}\xspace
  }%
}

\defterm\LTL{LTL}
\defterm\LTLP{LTL+P}
\defterm\CTL{CTL}
\defterm\ATL{ATL}
\defterm\EXPTIME{EXPTIME}

\newcommand\DeclareSymbol[2][\mathcal]{\csdef{#2}{\ensuremath{#1{#2}}\xspace}}

\DeclareSymbol[\mathbb]{N}
\DeclareSymbol[\mathbb]{Z}
\DeclareSymbol[\mathcal]{E}
\DeclareSymbol[\mathcal]{C}
\DeclareSymbol[\mathcal]{D}
\DeclareSymbol[\mathcal]{L}
\DeclareSymbol[\mathcal]{P}
\DeclareSymbol[\mathcal]{O}
\DeclareSymbol[\mathcal]{R}
\DeclareSymbol[\mathcal]{G}
\DeclareSymbol[\mathcal]{B}
\DeclareSymbol[\mathcal]{H}
\DeclareSymbol[\mathcal]{U}
\DeclareSymbol[\mathsf]{T}
\DeclareSymbol[\mathsf]{M}
\DeclareSymbol[\mathsf]{A}
\DeclareSymbol[\mathcal]{S}
\DeclareSymbol[\mathcal]{V}
\DeclareSymbol[\mathsf]{s}
\DeclareSymbol[\mathsf]{e}

\DeclareMathOperator\F{\mathsf{F}}

\renewcommand\implies{\longrightarrow}

\renewcommand\lor{\mathrel{\vee}}

\DeclarePairedDelimiter\seq{\langle}{\rangle}

\DeclarePairedDelimiter\abs{|}{|}

\newcommand\sE[1]{\langle\!\langle#1\rangle\!\rangle}

\newcommand\during{\mathbin{\mathsf{during}}}
\newcommand\meets{\mathbin{\mathsf{meets}}}
\newcommand\duration{\operatorname{\mathsf{duration}}}

\renewcommand\P{\mathcal{P}}

\DeclareDocumentCommand\letswap{m m}{%
  \let\temp#1
  \let#1#2
  \let#2\temp
  \let\temp\undefined
}

\letswap\emptyset\varnothing
\letswap\phi\varphi

\long\def\*#1*/{}

\newcommand\bfPi{\mathbf{\Pi}}

\renewcommand\bar[1]{\overline{#1}}

\newcommand\fitpar{\looseness=-1}

\newlang\PDDL{PDDL}
\newlang\STRIPS{STRIPS}
\newclass\NEXPSPACE{NEXPSPACE}
\newclass\NEXPTIME{NEXPTIME}
\newclass\PTIME{PTIME}

\newcommand\charlie{\textit{Charlie}\xspace}
\newcommand\eve{\textit{Eve}\xspace}

\newcommand\indam{%
  \raisebox{0.45ex}{\textexclamdown}%
  $\mathbb{N}\delta$%
  \kern-0.02em%
  A%
  \kern0.09em%
  \raisebox{-0.1ex}{\rotatebox[origin=c]{-90}{$\Sigma$}}%
  \kern0.09em%
  \xspace
}

\newcommand\platinum{\mbox{PLATINUm}\xspace}

\DeclareMathOperator\tokstart{\mathsf{start}}
\DeclareMathOperator\tokend{\mathsf{end}}
\DeclareMathOperator\play{\mathsf{play}}
\DeclareMathOperator\wait{\mathsf{wait}}
\DeclareMathOperator\now{\mathsf{now}}
\DeclareMathOperator\window{\mathsf{w}}

\newcommand\before[1][\le]{\ensuremath{#1}}

\DeclareDocumentCommand\DeclareTemporalRelation{m O{\le} m}{%
  \DeclareDocumentCommand#1{}{\before[#2]^{\mathsf{#3}}}%
}

\DeclareTemporalRelation\sbefores{s,s}
\DeclareTemporalRelation\ebeforee{e,e}
\DeclareTemporalRelation\sbeforee{s,e}
\DeclareTemporalRelation\ebefores{e,s}
\DeclareTemporalRelation\sbefore{s}
\DeclareTemporalRelation\ebefore{e}
\DeclareTemporalRelation\safter[\ge]{s}
\DeclareTemporalRelation\eafter[\ge]{e}
\DeclareTemporalRelation\startsat[=]{s}
\DeclareTemporalRelation\endsat[=]{e}

\DeclareMathOperator\starttime{\mathsf{start-time}}

\let\endtime\eendtime%

\newcommand\suchdot{\mathrel{.}}

\newcommand\true{\top}

\newcommand\SV{\ensuremath{\mathit{SV}}}

\def\timeline #1[#2] at #3 of length #4; {
  \coordinate (var#1) at #3;
  \path (var#1) node[left] {#2};
  \draw (var#1) -- +({#4},0) coordinate (endvar#1);
}

\def\tick at #1; {
  \draw ($#1+(0,\pad)$) -- ($#1-(0,\pad)$);
}
\def\midlabel #1[#2] between (#3) and (#4) #5; {
  \path (#3) -- (#4) node[midway,#5] (#1 label) {#2};
}

\tikzset{tokcolor/.style={gray}}

\newif\iftokenticks
\tokentickstrue
\def\token #1 at #2 of color #3, length #4, and label #5; {%
  \path #2 coordinate (s#1);
  \path (s#1) +(#4,0) coordinate (e#1);

  \fill[#3] ($(s#1) - (0,\pad/3)$) rectangle ($(e#1) + (0,\pad/3)$);
  \iftokenticks
    \tick at (s#1);
    \tick at (e#1);
  \fi
  \midlabel #1[#5] between (s#1) and (e#1) above;
}

\newif\ifpreprint
\preprinttrue

\title{%
  A game-theoretic approach to timeline-based planning with uncertainty%
}

\titlerunning{%
  A game-theoretic approach to timeline-based planning with uncertainty%
}

\author{Nicola Gigante \textnormal{and} Angelo Montanari}{%
  University of Udine, Udine, Italy%
}{gigante.nicola@spes.uniud.it\\angelo.montanari@uniud.it}{}{%
  N.\ Gigante and A.\ Montanari's work was mainly done while on leave at
  \emph{The University of Western Australia}. The work was supported by the GNCS project
  \emph{Formal methods for verification and synthesis of discrete and hybrid
  systems} (N.\ Gigante, A.\ Montanari, M.\ Cialdea Mayer) and the PRID project \emph{ENCASE - Efforts in the uNderstanding of
  Complex interActing SystEms} (N.\ Gigante, A.\ Montanari).%
}

\author{Marta Cialdea Mayer}{%
  University of Roma Tre,
  Rome, Italy
}{cialdea@ing.uniroma3.it}{}{}

\author{Andrea Orlandini}{%
  National Research Council,
  Rome, Italy
}{andrea.orlandini@istc.cnr.it}{}{}

\author{Mark Reynolds}{%
  The University of Western Australia,
  Perth, Australia
}{mark.reynolds@uwa.edu.au}{}{%
  Mark Reynolds acknowledges the support of Australian Research Council funding
  (DP140103365).
}

\authorrunning{%
  N. Gigante, A. Montanari, M. Cialdea Mayer, A. Orlandini, and M. Reynolds%
}

\Copyright{%
  N. Gigante, A. Montanari, M. Cialdea Mayer, A. Orlandini, and M. Reynolds%
}

\subjclass{%
  CCS → Computing methodologies → Artificial intelligence → Planning and scheduling → Planning under uncertainty%
}


\category{}

\relatedversion{}

\supplement{}

\funding{}

\acknowledgements{}

\EventEditors{Natasha Alechina, Kjetil N{\o}rv\r{a}g, and Wojciech Penczek}
\EventNoEds{3}
\EventLongTitle{%
  25th International Symposium on Temporal Representation and Reasoning (TIME 2018)%
}
\EventShortTitle{TIME 2018}
\EventAcronym{TIME}
\EventYear{2018}
\EventDate{October 15--17, 2018}
\EventLocation{Warsaw, Poland}
\EventLogo{}
\SeriesVolume{120}
\ArticleNo{13}
\nolinenumbers 

\ifpreprint
  \hideLIPIcs
\fi

\begin{document}

\maketitle

\begin{abstract}
  In timeline-based planning, domains are described as sets of independent, but
  interacting, components, whose behaviour over time (the set of timelines) is
  governed by a set of temporal constraints. A distinguishing feature of
  timeline-based planning systems is the ability to integrate planning with
  execution by synthesising control strategies for \emph{flexible plans}.
  However, flexible plans can only represent \emph{temporal uncertainty}, while
  more complex forms of nondeterminism are needed to deal with a wider range of
  realistic problems. In this paper, we propose a novel game-theoretic approach
  to timeline-based planning problems, generalising the state of the art while
  uniformly handling temporal uncertainty and nondeterminism. We define a
  general concept of timeline-based \emph{game} and we show that the notion of
  winning strategy for these games is strictly more general than that of control
  strategy for dynamically controllable flexible plans. Moreover, we show that
  the problem of establishing the existence of such winning strategies is
  decidable using a doubly exponential amount of space.
\end{abstract}

\section{Introduction}
\label{sec:introduction}

\raggedbottom

In the timeline-based approach to planning~\cite{Muscettola94}, the world is
modelled as a set of independent, but interacting, components, whose behaviour
over time, the \emph{timelines}, is governed by a set of temporal constraints.
This approach differs from the classical \emph{action-based} planning, relying
on \PDDL~\cite{FoxLong03}, for its more declarative nature and its focus on
temporal reasoning (see,~\eg~\cite{CialdeaMayerOU16,CimattiMR13,Frank13,GiganteMCO16,GiganteMCO17,Muscettola94}).
Timeline-based systems have been successfully deployed in a number of complex
scenarios, ranging from space operations
\cite{barreiro2012europa,cesta2009mrspock,Muscettola94} to manufacturing
\cite{BorgoCOU16,CestaMOU17}. An important feature of timeline-based planning
systems is their ability to integrate planning with execution by means of
\emph{flexible plans}, which represent envelopes of possible solutions that
differ in the execution times and/or the duration of tasks. Flexible plans allow
the controller to handle the \emph{temporal uncertainty} involved in dealing
with partially controllable elements and the external environment. Cialdea Mayer \etal \cite{CialdeaMayerOU16}
rigorously defined the concept of timeline-based planning specification as well as \emph{dynamically controllable} flexible
plans, which can be executed guaranteeing to satisfy the problem constraints
while reactively handling any \emph{temporal uncertainty} in the uncontrollable
behaviour. A technique for synthesising dynamic control strategies is shown in
\cite{CialdeaMayerO15}.

Other forms of uncertainty, such as \emph{nondeterminism} (\ie \emph{which}
tasks the environment chooses to perform), are not supported: even for external
variables, completely controlled by the environment, their evolution is known up
to temporal uncertainty only. This choice to focus on temporal reasoning and
temporal uncertainty is coherent with the history and scope of timeline-based
systems. However, it is not completely reflected into the grammar of modelling
languages used in timeline-based systems, which are expressive enough to model
complex scenarios that require the system to handle non-temporal nondeterminism.
In such cases, current systems often employ a \emph{re-planning} stage as part
of their execution cycle (see,~\eg~\cite{CestaMOU17}): any mismatch between the
expected and actual behaviours of the environment results into a revision of the
flexible plan, which then can resume execution. Unfortunately, the cost of such
a re-planning phase may be incompatible with the requirements of real-time
execution and, more importantly, if a wrong choice is made by the original
flexible plan, the re-planning might happen too late to be able to recover a
controllable state of the system. Hence, knowledge engineers have to explicitly
account for this problem if they want to avoid unnecessary failures and costly
re-planning during execution, which make the system less effective and more
complex to use.

Nondeterministic planning issues have been extensively investigated within the
action-based planning framework following different approaches such as, for
instance, reactive planning systems~\cite{beetzmcdermott94}, deductive
planning~\cite{steel94}, POMDP~\cite{kaelbling99}, and model
checking~\cite{cimatti03AIJ}. More recently, fully observable nondeterministic
(FOND) planning problems have been addressed~\cite{muise12,muise14} also
considering temporally extended goals~\cite{camacho16,patrizi13}. However,
action-based planning does not support flexible plans and temporal uncertainty,
and it does not take into account controllability issues. Recently, SMT-based
techniques have been exploited to deal with uncontrollable durations in strong
temporal planning~\cite{cimatti18}; however, dynamic controllability issues are
not addressed.\fitpar

This paper defines the novel concept of \emph{timeline-based planning game}, a
game-theoretic generalisation of the timeline-based planning problem with
uncertainty, which  uniformly treats both temporal uncertainty and general
nondeterminism. In these games, the controller tries to satisfy the given
temporal constraints no matter what the choices of the environment are. We
compare the proposed games with the current approaches based on flexible plans.
In particular, we show how current timeline-based modelling languages can express
problems that, only seeming to involve temporal uncertainty at first, in fact
model scenarios which would require the controller to handle non-temporal
nondeterminism. We show that these problems do not admit dynamically
controllable flexible plans (as defined in \cite{CialdeaMayerOU16}), but do
admit winning strategies when seen as instances of timeline-based games.
A study of the decidability and complexity of the problem of establishing the existence
of a winning strategy for a given timeline-based planning game concludes
the paper.

The paper is structured as follows. \Cref{sec:timelines} briefly recaps the
basic definitions, and \cref{sec:issues} discusses the limitations of the
approach employed in state-of-the-art timeline-based planning systems. Then,
\cref{sec:game} defines timeline-based planning games, and shows its greater
generality with respect to the current approach. Finally,
\Cref{sec:decidability} addresses decidability and complexity issues.
\Cref{sec:conclusions} concludes discussing future developments.

\flushbottom

\section{Timeline-based planning}
\label{sec:timelines}

This section introduces timeline-based planning and describes the state of the
art of the field with regards to how uncertainty is handled by current
timeline-based systems. As a representative of the modelling languages used by
existing systems, we chose the formal language introduced in
\cite{CialdeaMayerOU16}. We first introduce the basic concepts of the framework,
without considering uncertainty, as studied in \cite{GiganteMCO16,GiganteMCO17};
then, we add uncertainty to the picture and discuss how it is handled by
current systems.

\subsection{Basic definitions}
State variables are the basic building blocks of the timeline-based planning
framework.

\begin{definition}[State variables]
  \label{def:statevar}
  A \emph{state variable} is a tuple $x=(V_x,T_x,D_x,\gamma_x)$, where:
  \begin{itemize}
  \item $V_x$ is the \emph{finite domain} of the variable;
  \item $T_x:V_x\to2^{V_x}$ is the \emph{value transition function}, which maps
        each value $v\in V_x$ to the set of values that can follow it;
  \item $D_x:V_x\to\N^+\times(\N^+\cup\{+\infty\})$ is a function that maps each
        $v\in V_x$ to the pair $(d^{x=v}_{min},d^{x=v}_{max})$ of the minimum
        and maximum duration of any interval over which $x=v$;
  \item $\gamma_x:V_x\to\{\mathsf{c},\mathsf{u}\}$ is a function called
        \emph{controllability tag} (see \cref{subsec:flexible-timelines}).
  \end{itemize}
\end{definition}

Which value is taken by a state variable over a specified time interval is
described by means of \emph{tokens}. The behaviour of a state variable over
time is modelled by a finite sequence of tokens, called a \emph{timeline}.

\begin{definition}[Timelines]
  \label{def:timeline}
  A \emph{token} for $x$ is a tuple $\tau=(x,v,d)$, where $x$ is a state
  variable, $v\in V_x$ is the value held by the variable, and $d\in\N^+$ is the
  \emph{duration} of the token. A \emph{timeline} for a state variable
  $x=(V_x,T_x,D_x,\gamma_x)$ is a finite sequence
  $\T=\seq{\tau_1,\ldots,\tau_k}$ of tokens for $x$.
\end{definition}

For any token $\tau_i=(x,v_i,d_i)$ in a timeline $\T =
\seq{\tau_1,\ldots,\tau_k}$ we can define the functions $\starttime(\T,i) =
\sum_{j=1}^{i-1} d_j$ and $\endtime(\T,i) = \starttime(\T,i) + d_i$, hence
mapping each token to the corresponding time interval $[\starttime,\endtime)$
(right extremum excluded). As an example, the time interval associated with the
token $\tau_1=(x,2,5)$ is $[0,5)$. When there is no ambiguity, we write
$\starttime(\tau_i)$ and $\endtime(\tau_i)$ to denote, respectively,
$\starttime(\T,i)$ and $\endtime(\T,i)$. The \emph{horizon} of a timeline
$\T=\seq{\tau_1,\ldots,\tau_k}$ is defined as $\H(\T)=\endtime(\tau_k)$. A
timeline $\T$ can be empty, in which case we define its horizon as $\H(\T)=0$.

The problem domain and the goal are modelled by a set of temporal constraints,
called \emph{synchronisation rules}. For the sake of space, we do not provide a
detailed account of their syntax. Informally, each synchronisation rule has the
following form:
\begin{align*}
  \mathit{rule} & {} \mathrel{:=}
    a_0[x_0=v_0]\implies \E_1\lor\E_2\lor\ldots\lor\E_k, \ with \\
  \E_i & {} \mathrel{:=} \exists
    a_1[x_1=v_1]a_2[x_2=v_2]\ldots a_n[x_n=v_n]\suchdot \C
\end{align*}
where $x_0, \ldots, x_n$ are state variables and $v_0, \ldots, v_n$ are values,
with $v_i\in V_{x_i}$ for all $i$. Each rule thus consists of a \emph{trigger}
($a[x_0=v_0]$) and a disjunction of \emph{existential statements}. It is
satisfied if for each token satisfying the trigger, at least one of the
disjuncts is satisfied. The trigger can be empty ($\T$), in which case the rule
is said to be \emph{triggerless} and asks for the satisfaction of the body
without any precondition. Each existential statement requires the existence of
some tokens such that the clause $\C$ is satisfied. The clause is in turn a
conjunction of \emph{atoms}, that is, atomic relations between endpoints of the
quantified tokens, of the form $a\le^{e_1,e_2}_{[l,u]} b$, where $a$ and $b$ are
token names, $e_1,e_2\in\{\mathsf{s},\mathsf{e}\}$, $l\in\N$, and
$u\in\N\cup\{+\infty\}$. Intuitively, each atom relates the start ($\mathsf{s}$)
or end ($\mathsf{e}$) of the two tokens, and $l$ and $u$ are respectively a
lower and upper bound to the distance between the two endpoints. Pointwise atoms
relating a single endpoint with a specific point in time are also possible, \eg
$a\le^{\mathsf{s}}_{[l,u]} t$. An atom $a\le^{e_1,e_2}_{[l,u]}b$ is
\emph{bounded} if $u\ne+\infty$, and \emph{unbounded} otherwise. With these
basic atomic relations, one can express all the Allen's relations over time
intervals~\cite{Allen83}. As an example, one can define $a\meets b$ as
$a\ebefores_{[0,0]} b$, or $a\during b$ as $a\sbefores_{[0,+\infty]} b\land b
\ebeforee_{[0,+\infty]} a$. Moreover, one can constrain the duration of tokens,
\eg writing $\duration(a)=k$ or $\duration(a)\ge k$ as a shorthand for,
respectively, $a\sbeforee_{[k,k]} a$ and $a\sbeforee_{[k,+\infty]} a$.
Moreover, disjunctions in synchronisation rules allows one to express some forms
of conditional (if/then/else) statements.

In the simplest setting, a timeline-based planning problem consists of a set of
state variables and a set of rules that represent the problem domain and the
goal. A solution to such a problem is simply a set of timelines that satisfy the
rules.

\begin{definition}[Timeline-based planning problem]
  \label{def:problem}
  A \emph{timeline-based planning problem} is a pair $P=(\SV,S)$, where
  $\SV$ is a set of state variables and $S$ is a set of synchronisation rules
  over $\SV$.
\end{definition}
\begin{definition}[Solution plan]
  \label{def:plan}
  A \emph{scheduled solution plan} for a problem $P=(\SV,S)$ is a set $\Gamma$
  of scheduled timelines, one for each $x\in\SV$, such that $v_{i+1}\in
  T_x(v_i)$ and $d_{min}^{x=v}\le d_i \le d_{max}^{x=v}$ for all tokens
  $\tau_i=(x_i,v_i,d_i)\in\T_x$, and all the rules in $S$ are satisfied.
\end{definition}

Note that the controllability tags $\gamma_x$ of state variables are ignored in
Definition \ref{def:plan}, as it only considers the problem of satisfying the
synchronisation rules and the issues arising from the execution of the plan are
not considered. Even without considering any form of uncertainty, timeline-based
planning is already quite a hard problem. Indeed, we know from
\cite{GiganteMCO17} that the problem of deciding whether there exists a solution
plan for a given timeline-based planning problem is \EXPSPACE-complete.

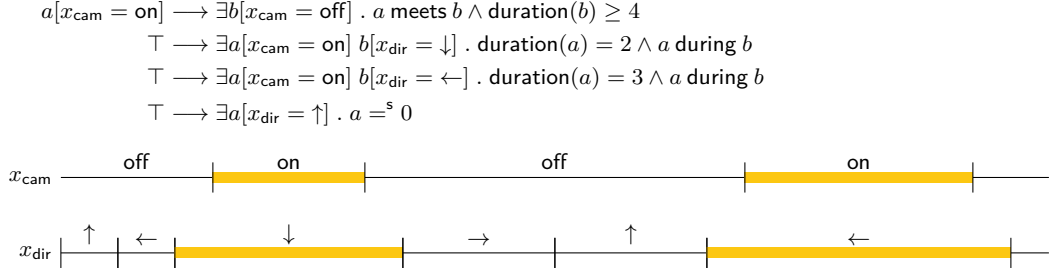
\begin{figure}

  \footnotesize

  \begin{align*}
    a[x_{\mathsf{cam}}=\mathsf{on}]\implies {} & \exists
      b[x_\mathsf{cam}=\mathsf{off}] \suchdot a \meets b \land \duration(b)\ge 4\\
    \true \implies {} & \exists
      a[x_{\mathsf{cam}}=\mathsf{on}] \; b[x_{\mathsf{dir}}=\;\downarrow]
        \suchdot \duration(a)=2 \land a \during b\\
    \true \implies {} & \exists
      a[x_{\mathsf{cam}}=\mathsf{on}] \; b[x_{\mathsf{dir}}=\;\leftarrow]
        \suchdot \duration(a)=3 \land a \during b \\
    \true \implies {} & \exists a[x_{\mathsf{dir}}=\;\uparrow]\suchdot
      a =^{\mathsf{s}} 0
  \end{align*}

  \begin{tikzpicture}[every node/.append style={math mode}]
    \def\h{3cm}

    \def\pad{0.2cm}

    \def\w{\textwidth-5*\pad}

    \coordinate (tx) at ({2*\pad},{2/3*\h-\pad});
    \coordinate (ty) at ({2*\pad},{\h/3-\pad});

    \timeline x[x_{\mathsf{cam}}] at (tx) of length \w;
    \timeline y[x_{\mathsf{dir}}] at (ty) of length \w;

    \token a at (varx) +(2cm,0) of color lipics, length 2cm,
      and label {\mathsf{on}};

    \token b at (ea) +(5cm,0) of color lipics, length 3cm,
      and label {\mathsf{on}};

    \token dir1 at (vary) of color transparent, length 0.75cm,
      and label {\uparrow};

    \token dir2 at (edir1) of color transparent, length 0.75cm,
      and label {\leftarrow};

    \token dir3 at (edir2) of color lipics, length 3cm,
      and label {\downarrow};

    \token dir4 at (edir3) of color transparent, length 2cm,
      and label {\rightarrow};

    \token dir5 at (edir4) of color transparent, length 2cm,
      and label {\uparrow};

    \token dir6 at (edir5) of color lipics, length 4cm,
      and label {\leftarrow};

    \midlabel off[\mathsf{off}] between (varx) and (sa) above;
    \midlabel off[\mathsf{off}] between (ea) and (sb) above;

  \end{tikzpicture}

  \caption{%
    An example of timeline-based planning problem. Two state variables are used
    to represent the on/off state of a camera $x_{\mathsf{cam}}$ and its
    pointing direction $x_{\mathsf{dir}}$. The transition function of $x_{dir}$
    forces the camera to only move counterclockwise.%
  }
  \label{ex:rules}
\end{figure}

As an example, consider \cref{ex:rules}, which shows a possible solution for a
problem with two state variables, $x_{\mathsf{cam}}$ and $x_{\mathsf{dir}}$,
that respectively represent the on/off state of a camera and its direction. The
transition function $T_{x_{\mathsf{dir}}}$ of the second variable is such that
the camera can only stay still or move counterclockwise, that is,
$T_{x_{\mathsf{dir}}}(\leftarrow)=\{\leftarrow,\downarrow\}$. The first rule
states the system requirement that the camera must remain switched off at least
four time steps after each use, to let the components cool down. The objective
of the system is that of performing some shoots of a given duration, provided
the camera is pointed in the right direction: a shoot downwards lasting two time
steps, and a shoot toward left lasting three time steps, in an arbitrary order.
This objective is encoded by the second and third synchronisation rules, which
are triggerless. The last rule expresses the fact that the camera is initially
pointed upwards.\fitpar

\subsection{Timeline-based planning with uncertainty}
\label{subsec:flexible-timelines}

A distinguishing feature of existing timeline-based planning systems is their
ability to integrate planning and execution, accounting for the unavoidable
uncertainty that comes from the interaction with the environment where the plan
is executed. We now recall how timeline-based planning problems with uncertainty
are defined in \cite{CialdeaMayerOU16}.

Two different sources of uncertainty can be represented by this model. The first
comes from \emph{external variables}, which are completely under the control of
the environment, in contrast to \emph{controlled variables} (also called
\emph{planned} variables), which are dealt with the system. The planner is not
allowed to decide anything regarding the behaviour of external variables. The
second is the duration of \emph{uncontrollable tokens}. The
\emph{controllability tag} $\gamma_x$, associated with each state variable $x$,
states whether a token where $x=v$ is \emph{controllable}
($\gamma_x(v)=\mathsf{c}$) or \emph{uncontrollable} ($\gamma_x(v)=\mathsf{u})$.
The execution of uncontrollable tokens is planned and performed by the system,
but their duration cannot be known in advance, \eg one may go shopping without
knowing how much time he/she will have to wait at the counter. In both cases,
only \emph{temporal} uncertainty is considered in the current approach.\fitpar

To deal with the uncertainty inherent in the execution of the plan,
timeline-based planning systems make use of the concept of \emph{flexible
timeline}.

\begin{definition}[Flexible timeline]
  \label{def:flexible-timeline}
  A \emph{flexible token} is a tuple $\tau=(x,v,[e,e'],[d,d'])$, where $x$ is a
  state variable, $v\in V_x$, $[e,e']\in\N\times\N$ is the interval of possible
  token \emph{end times}\footnotemark, and $[d,d']\in\N^+\times\N^+$ is the
  interval of possible token \emph{durations}.

  A \emph{flexible timeline} for a state variable $x=(V_x,T_x,D_x,\gamma_x)$ is
  a finite sequence $\T_x=\seq{\tau_1,\ldots,\tau_k}$ of flexible tokens for $x$
  where $[e_1,e_1']=[d_1,d_1']$, and
  $[e_i,e_i']\subseteq[e_{i-1}+d_i,e_{i-1}'+d_i']$ for all $1 < i \le k$.
\end{definition}

\footnotetext{%
  Flexible tokens report their end times rather than their start times because
  in this way a flexible timeline can precisely constrain its horizon.%
}

A flexible timeline represents a set of different timelines which differ in the
precise timings of the described events. Tokens, timelines, and plans will also
be referred as \emph{scheduled} tokens, timelines and plans, to better
differentiate them from flexible ones. A scheduled timeline
$\T=\seq{\tau_1,\ldots,\tau_k}$ is an \emph{instance} of a flexible timeline
$\T'=\seq{\tau'_1,\ldots,\tau'_k}$ if for each $\tau_i=(x,v,d)$ and
$\tau'_i=(x',v',[e,e'],[d,d'])$, we have $x=x'$, $v=v'$, $d\in[d,d']$, and
$\endtime(\tau_i)\in[e,e']$.

Flexibility can be naturally lifted from timelines to plans. The notion of
\emph{flexible plan} is formally defined as follows.
\begin{definition}[Flexible plan]
  \label{def:flexible-plan}
  A \emph{flexible plan} over a set of state variables $\SV$ is a pair
  $\Pi=(\Gamma,\R)$, where $\Gamma$ is a set of \emph{flexible timelines},
  exactly one for each $x\in\SV$, and $\R$ is a set of \emph{atoms} over the
  tokens occurring in $\Gamma$.
\end{definition}

Given a flexible plan $\Pi=(\Gamma,\R)$, a scheduled plan
$\Gamma'$ is an \emph{instance} of $\Pi$ if the timelines in $\Gamma'$ are
instances of those in $\Gamma$, and they satisfy the atoms in $\R$.

A flexible plan can be viewed as a tentative set of solutions to a planning
problem where the precise timing of execution and the duration of tokens are
chosen during execution. The set of atoms that comes together with the set of
flexible timelines allow the plan to specify additional constraints over the
tokens that compose the timelines. Flexible plans can in particular be used to
describe the expected behaviour of external variables, of which one may know the
future evolution only up to some temporal uncertainty. It is worth pointing out
that a flexible plan is \emph{unconditional}, \ie a single plan is committed to
a specific sequence of state variable values, and the only freedom left concerns
token durations.

A timeline-based planning problem with uncertainty is formally defined as
follows.
\begin{definition}[Timeline-based planning problem with uncertainty]
  \label{def:problem:uncertainty}
  A \emph{timeline-based planning problem with uncertainty} is a tuple
  $P=(\SV_C,\SV_E,S,\O)$, where:
  \begin{itemize}
  \item $\SV_C$ and $\SV_E$ are the sets of, respectively, the
        \emph{controlled} and the \emph{external} variables;
  \item $S$ is a set of synchronisation rules over $\SV_C\cup\SV_E$;
  \item the \emph{observation} $\O=(\Gamma_E,\R_E)$ is a flexible plan over
        $\SV_E$ specifying the behaviour of external variables.
  \end{itemize}
\end{definition}

Definition \ref{def:problem:uncertainty} differs from Definition
\ref{def:problem} in two main respects: it splits the set of variables into
controlled and external ones and it includes a flexible plan describing the
temporally uncertain behaviour of external variables.

To solve the problem, one has to find a set of flexible timelines for the
controlled variables such that the rules can be satisfied by a suitable set of
instances.

\begin{definition}[Flexible solution plan]
  \label{def:flexible-solution-plan}
  Let $P=(\SV_C,\SV_E,S,\O)$, with $\O=(\Gamma_E,\R_E)$ be a timeline-based planning
  problem with uncertainty. A flexible plan $\Pi=(\Gamma,\R)$ over $\SV_C\cup\SV_E$
  is a \emph{flexible solution plan} for $P$ if:
  \begin{enumerate}
  \item \label{def:flexible-solution-plan:observation}
        $\Pi$ agrees with $\O$, that is, $\Gamma_E\subseteq\Gamma$ and
        $\R_E\subseteq\R$;
  \item \label{def:flexible-solution-plan:no-restriction}
        the plan does not restrict the duration of uncontrollable tokens, that is,
        for any $\T\in\Gamma$ and any token $\tau=(x,v,[e,e'],[d,d'])\in\T$, if
        $\gamma_x(v)=\mathsf{u}$, then $d=d^{x=v}_{min}$ and $d'=d^{x=v}_{max}$;
  \item \label{def:flexible-solution-plan:instances}
        any \emph{instance} of $\Gamma$ is a scheduled solution plan for the
        timeline-based planning problem $P'=(\SV_C\cup\SV_E,S)$, and there
        exists at least one such instance.
  \end{enumerate}
\end{definition}

Note that, despite the name, which is borrowed from \cite{CialdeaMayerOU16}, the
observation $\O$ is rather an \apriori description of the environment behaviour,
which is supposed to be completely known up to the given temporal uncertainty.
Usual definitions of planning problems involve the specification of a maximum
bound on the \emph{horizon} of the solution plans. For the sake of generality,
we omit this parameter, as it can be expressed by suitable synchronisation
rules.

According to \cref{def:flexible-solution-plan:instances} of
Definition~\ref{def:flexible-solution-plan}, any instance of a flexible plan
satisfies the synchronisation rules of the problem. However, there is no
guarantee that one such instance exists for each possible instance of the
external timelines. In other words, Definition~\ref{def:flexible-solution-plan}
does not guarantee that a flexible solution plan can be executed in any possible
scenario. Thus, a control strategy is needed to determine how to schedule
controllable tasks.
Because of space concerns, we cannot present all the details of the definition
of control strategy, which is thoroughly illustrated in \cite{CialdeaMayerOU16}.
Informally, it can be thought of as a function $\sigma$ which chooses how to
schedule the start time of tokens for controlled variables, and the end time of
controllable tokens, during execution.
A flexible solution plan $\Pi$ is said to be \emph{weakly controllable} if for
each possible schedule of tokens for external variables and of uncontrollable
tokens, there is a control strategy $\sigma$ such that following $\sigma$
during execution results into an instance of $\Pi$.
It is said to be \emph{strongly controllable} if, conversely, a single control
strategy $\sigma$ exists which results into an instance of $\Pi$ whatever
the schedule of the endpoints under its control by the environment is.
Finally, it is said to be \emph{dynamically controllable} if the controlled
endpoints can be scheduled by taking into account, at any given time, only
the past history of the execution in such a way that an instance of $\Pi$
is obtained.
Given its generality and wider applicability, dynamic controllability is
definitely the most interesting form of controllability.
Notice that these concepts, whose definition in the context of flexible plans
for timeline-based planning is given in \cite{CialdeaMayerOU16}, have analogous
counterparts in the context of temporal networks~\cite{VidalF99}.\fitpar

Timeline-based systems which aim at handling both planning and execution cannot
simply produce flexible plans, but have to ensure a chosen degree of
controllability of the produced plans.
As an example, the \platinum system~\cite{CestaMOU17} employs a two-phase
process where a flexible plan is first produced and then checked for the
existence of a dynamic control strategy. Dynamic controllability of a flexible
plan can be checked, for instance, via a reduction to timed game automata
\cite{CialdeaMayerO15}. Since uncontrollable flexible plans are not suitable to
be executed, the problem is considered to be solved only when a dynamically
controllable flexible solution plan is found, together with its dynamic control
strategy.

\section{Limitations of the current approach}
\label{sec:issues}

The focus of timeline-based systems on temporal reasoning and temporal
uncertainty clearly emerges from the previous section. This focus has its roots
in the history of the paradigm and the typical application scenarios where
timeline-based systems have been employed. The exclusive focus on temporal
uncertainty is especially evident in the treatment of external variables: their
behaviour is supposed to be completely known excepting only the precise timing
of specific events. Consider, for example, a satellite in orbit doing some
measurements and transmitting the results back to Earth. In such a domain,
external variables might be used to represent visibility windows where the
different ground stations can be reached by the satellite. The precise timing of
those windows is uncertain, but everything else is known (even months) in
advance. Nevertheless, to handle the case of a mismatch between the expected and
the observed behaviour of the environment, systems such as \platinum employ a
feedback loop where a \emph{failure manager} is in charge of triggering, if
needed, a \emph{re-planning} (or plan repair) phase, which should produce a new
flexible plan, with a suitable control strategy, taking into account the newly
acquired observations (the name of the $\O$ component in
Definition~\ref{def:problem:uncertainty} comes from this scenario).\fitpar

In contrast to domains as the above one, other applications might require to
re-plan more frequently. As an example, in robotics scenarios such as those
discussed in \cite{CestaMOU17}, where the planned system interacts with a human
agent, one cannot hope to represent with temporal uncertainty alone all the
possible variability of the behaviour of the external environment. In these
scenarios, most often re-plans get triggered to handle the unpredicted outcomes
of generally nondeterministic choices that the external agents can make, rather
than to fix problems in the domain model. This observation motivates us in
proposing an interpretation of timeline-based problems that is able to handle
both temporal uncertainty and general nondeterminism, extending and generalising
the current approach based on flexible plans.\fitpar

As a matter of fact, it turns out that the domain description languages commonly
in use, here exemplified by the formal syntax defined in \cref{sec:timelines},
can easily express scenarios where the need to handle general nondeterminism
arise in problems which apparently only involved temporal uncertainty. To see
how this may happen, consider a simple timeline-based planning problem with
uncertainty $P=(\SV_C,\SV_E,S,\O)$, with a single state variable $x$, with
$V_x=\{v_1,v_2,v_3\}$, $\SV_E = \emptyset$, and $S$ consisting of the following
synchronisation rules:
\begin{align*}
  a[x=v_1] \implies {} & \exists b[x=v_2] \suchdot a \ebefores_{[0,0]} b
                                             \land a \sbeforee_{[0,5]} a 
           {} \lor  {}
\exists c[x=v_3] \suchdot a \ebefores_{[0,0]} c
                                             \land a \sbeforee_{[6,10]} a\\
  \true     \implies {} & \exists a[x=v_1] \suchdot a \startsat 0
\end{align*}

Suppose that $D_x(v)=[1,10]$ for all $v\in V_x$, and that
$\gamma_x(v_1)=\mathsf{u}$ and $\gamma_x(v_2)=\gamma_x(v_3)=\mathsf{c}$, that
is, tokens where $x=v_1$ are uncontrollable. The rules require the execution to
start with a token where $x=v_1$, followed by a token where either $x=v_2$ or
$x=v_3$ depending on the duration of the first token. This scenario is,
intuitively, trivial to control. The system must execute $x=v_1$ as a first
token due to the second rule. Then, the environment controls its duration, and
the system simply has to wait for the token to end, and then execute either
$x=v_2$ or $x=v_3$ depending on how long the first token lasted. However, there
are no flexible plans that represent this simple strategy, since each given plan
must fix the value of every token in advance. To guarantee the satisfaction of
the rules, the choice of which value to assign to $x$ on the second token must
be made during the execution, but this is not possible because of the
exclusively sequential nature of flexible plans. In this case, therefore, the
problem would be considered as unsolvable, even if the goals stated by the rules
seem simple to achieve.\fitpar

The issues discussed here come from the lack of a proper support for general
nondeterminism in the framework of flexible plans. However, the last example
shows  that this is not just a missing feature of current systems, but rather a
class of scenarios that can be easily modelled by (the syntax of) timeline-based
description languages but whose solutions are not captured by the commonly
considered semantics. The \emph{timeline-based planning games} defined in the
next section provide a clean way to express the solution to this kind of
scenario, providing a semantics to timeline-based planning problems with
uncertainty capable of modelling both temporal uncertainty and general
nondeterminism in a uniform way. Moreover, they handle the external variables in
the most general way, without assuming any \apriori knowledge of their future
behaviour.

\section{Timeline-based planning games}
\label{sec:game}
Let us now introduce \emph{timeline-based planning games}. They generalise
dynamic control strategies for flexible plans while suitably handling the
limitations discussed in \cref{sec:issues}.

\begin{definition}[Timeline-based planning game]
  \label{def:game}
  A \emph{timeline-based planning game} is a tuple $G=(\SV_C,\SV_E,\S,\D)$,
  where $\SV_C$ and $\SV_E$ are the sets of, respectively, the \emph{controlled}
  and the \emph{external} variables, and $\S$ and $\D$ are two sets of
  synchronisation rules, respectively called \emph{system}  and \emph{domain}
  rules, involving variables from both $\SV_C$ and $\SV_E$.
\end{definition}

Intuitively, a timeline-based planning game $G=(\SV_C,\SV_E,\S,\D)$ is a
turn-based, two-player game played by the controller, \charlie, and the
environment, \eve. By playing the game, the players progressively build the
timelines of a \emph{scheduled plan} (see Definition~\ref{def:plan}). At each
round, each player makes a move deciding which tokens to start and/or to stop
and at which time. Both players are constrained by the set $\D$ of \emph{domain}
rules, which describe the basic rules governing the world.  Domain rules replace
and generalise the \emph{observation} $\O$ of
Definition~\ref{def:problem:uncertainty}, allowing one to freely model the
interaction between the system and the environment. They are not intended to be
\eve's (or \charlie's) \emph{goals}, but, rather, a background knowledge about
the world that can be assumed to hold at any time. Since neither player can
violate $\D$, the strategy of each player may safely assume the validity of such
rules. In addition, \charlie is responsible for satisfying the set $\S$ of
\emph{system} rules, which describe the rules governing the controlled system,
including its goals. \charlie wins if, assuming \eve behaves according to the
domain rules, he manages to construct a plan satisfying the system rules. In
contrast, \eve wins if, while satisfying the domain rules, she prevents \charlie
from winning, either by forcing him to violate some system rule, or by
indefinitely postponing the fulfilment of his goals.\fitpar

Let us now formally describe the way in which a play of a (timeline-based)
planning game evolves. First of all, we observe that at any given time during
the play, the plan will be partially built, waiting for some tokens to be
completed. A \emph{partial plan} is a plan where the last token on each timeline
may be unfinished (\emph{open} token). A timeline whose last token is open is
called an \emph{open} timeline.

\begin{definition}[Open timeline]
  \label{def:opentimelines}
  Let $G=(SV_C,SV_E,\S,\D)$ be a planning game and let $SV = SV_C\, \cup\,
  SV_E$. An \emph{open token} for $G$ is a pair $\tau=(x,v)$, where $x\in\SV$
  and $v\in V_x$. An \emph{open timeline} for $x\in\SV$ is a non-empty finite
  sequence of tokens $\T=\seq{\tau_1,\ldots,\tau_{k-1},\tau_k}$, where
  $\seq{\tau_1,\ldots,\tau_{k-1}}$ is a scheduled timeline for $x$ and
  $\tau_k=(x,v_k)$ is an open token.
\end{definition}

We will refer to tokens and timelines as defined in
Definition~\ref{def:timeline} as \emph{closed} tokens and \emph{closed}
timelines, respectively. In an open timeline $\T=\seq{\tau_1,\ldots,\tau_k}$,
only $\starttime(\tau_k)$ is defined for its last open token $\tau_k$:
$\starttime(\tau_0)=0$ and $\starttime(\tau_i)=\endtime(\tau_{i-1})$ for $i>1$.
Recall that $\H(\T)$ for a closed timeline is the $\endtime$ of its last token
and that $\H(\T)=0$ for empty timelines. For an open timeline
$\T=\seq{\tau_1,\ldots,\tau_{k-1},\tau_k}$, we define its horizon as the
$\endtime$ of its last \emph{closed} token, \ie $\H(\T)=\endtime(\tau_{k-1})$
(which is equal to $\starttime(\tau_k)$).\fitpar

\begin{definition}[Partial plan]
  \label{def:partialplan}
  Let $G=(SV_C,SV_E,\S,\D)$ be a planning game and let $SV = SV_C \ \cup \
  SV_E$.  A \emph{partial plan} for $G$ is a pair $\Pi=(\Gamma, \now)$, where
  $\Gamma$ is a set of timelines, either open or closed, one for each $x\in\SV$,
  and $\now\in\N$ is the current time, such that:
  \begin{enumerate}
  \item \label{def:partialplan:horizon}
        $\H(\T) < \now$ for any \emph{open} timeline $\T\in\Gamma$;
  \item \label{def:partialplan:synch}
        $\H(\T)=\now$ for any \emph{closed} timeline $\T\in\Gamma$;
  \end{enumerate}
\end{definition}

\begin{figure}

  \begin{tikzpicture}[xscale=0.9,yscale=1.2,every node/.append style={math mode}]

    \def\h{3cm}
    \def\pad{0.2cm}
    \def\w{12cm}

    \begin{scope}
      \node (a) at (-0.5cm,-0.5cm) {\text{(a)}};
      \coordinate (x) at (0,0);
      \coordinate (y) at (0,-1cm);

      \draw (x) node[left] {x} -- ++(3.2cm,0) node (endx) {};
      \draw (y) node[left] {y} -- ++(3.2cm,0) node (endy) {};
      \draw[dashed] (endx) -- ++(0.8cm,0);
      \draw[dashed] (endy) -- ++(0.8cm,0);

      \foreach \x in {1,3} {
        \draw (\x,0) -- ++(0,-\pad);
        \draw (\x,-1cm) -- ++(0,-\pad);
      }

      \foreach \x in {0,1,3} {
        \draw (\x,-\pad) node[below] {\x};
        \draw (\x,-1cm-\pad) node[below] {\x};
      }

      \draw (1.8,-\pad) node[below] {2};
      \draw (1.8,-1cm-\pad) node[below] {2};

      \token x1 at (x) of color lipics, length 2cm, and label {\tau_1};
      \token y1 at (y) of color lipics, length 2cm, and label {\tau_2};

      \draw[dashed] ($(ex1)+(0,0.5cm)$) -- ($(ey1)+(0,-0.8cm)$)
           node[below] {\mathsf{now}};
    \end{scope}

    \begin{scope}[xshift=5cm]
      \node (b) at (-0.5cm,-0.5cm) {\text{(b)}};
      \coordinate (x) at (0,0);
      \coordinate (y) at (0,-1cm);

      \draw (x) node[left] {x} -- ++(3.2cm,0) node (endx) {};
      \draw (y) node[left] {y} -- ++(3.2cm,0) node (endy) {};
      \draw[dashed] (endx) -- ++(0.8cm,0);
      \draw[dashed] (endy) -- ++(0.8cm,0);

      \foreach \x in {1,3} {
        \draw (\x,0) node (tick\x) {} -- ++(0,-\pad);
      }

      \foreach \x in {2,3} {
        \draw (\x,-1cm) node (tick\x) {} -- ++(0,-\pad);
      }

      \foreach \x in {0,1,3} {
        \draw (\x,-\pad) node[below] {\x};
        \draw (\x,-1cm-\pad) node[below] {\x};
      }
      \draw (1.8,-\pad) node[below] {2};
      \draw (1.8,-1cm-\pad) node[below] {2};

      \token x1 at (x) of color lipics, length 2cm, and label {\tau_1};
      \token y1 at (y) of color lipics, length 1cm, and label {\tau_2};

      \tokenticksfalse
      \token y2 at (y) +(1cm,0) of color gray, length 1cm, and label {\tau_3};
      \token y2' at (y) +(2cm,0)
              of color {gray, pattern=north west lines, pattern color=gray},
              length 2cm, and label {};

      \draw[dashed] ($(ex1)+(0,0.5cm)$) -- ($(ey2)+(0,-0.8cm)$)
           node[below] {\mathsf{now}};
    \end{scope}

    \begin{scope}[xshift=10cm]
      \node (c) at (-0.5cm,-0.5cm) {\text{(c)}};
      \coordinate (x) at (0,0);
      \coordinate (y) at (0,-1cm);

      \draw (x) node[left] {x} -- ++(3.2cm,0) node (endx) {};
      \draw (y) node[left] {y} -- ++(3.2cm,0) node (endy) {};
      \draw[dashed] (endx) -- ++(0.8cm,0);
      \draw[dashed] (endy) -- ++(0.8cm,0);

      \foreach \x in {1,3} {
        \draw (\x,0) node (tick\x) {} -- ++(0,-\pad);
      }

      \foreach \x in {2,3} {
        \draw (\x,-1cm) node (tick\x) {} -- ++(0,-\pad);
      }

      \foreach \x in {0,1,2} {
        \draw (\x,-\pad) node[below] {\x};
        \draw (\x,-1cm-\pad) node[below] {\x};
      }

      \draw (2.8,-\pad) node[below] {3};
      \draw (2.8,-1cm-\pad) node[below] {3};

      \token x1 at (x) of color lipics, length 2cm, and label {\tau_1};
      \token y1 at (y) of color lipics, length 1cm, and label {\tau_2};

      \tokenticksfalse
      \token y2 at (y) +(1cm,0) of color gray, length 2cm, and label {\tau_3};
      \draw ($(ey2)+(0,\pad)$) -- ($(ey2)+(0,-\pad)$);

      \draw[dashed] ($(ex1)+(1cm,0.5cm)$) -- ($(ey2)+(0,-0.8cm)$)
           node[below] {\mathsf{now}};

      \begin{scope}[line width=0.2cm, red, opacity=0.2, line cap=round]
        \draw (0,0.5cm) -- (4cm,-2cm);
        \draw (0,-2cm) -- (4cm,0.5cm);
      \end{scope}
    \end{scope}
  \end{tikzpicture}

  \caption{%
    Examples of partial plans: (a) two tokens for variables $x$ and $y$ where
    $\endtime(\tau_1)=\endtime(\tau_2)=2$ and $\now=2$; (b)
    $\endtime(\tau_1)=2$, $\endtime(\tau_2)=1$, while $\tau_3$ continues; (c)
    incorrect partial plan, because $\tau_3$ continues over the end of $\tau_1$
    and a successor for $\tau_1$ is unspecified.%
  }
  \label{fig:partial-plans}
\end{figure}
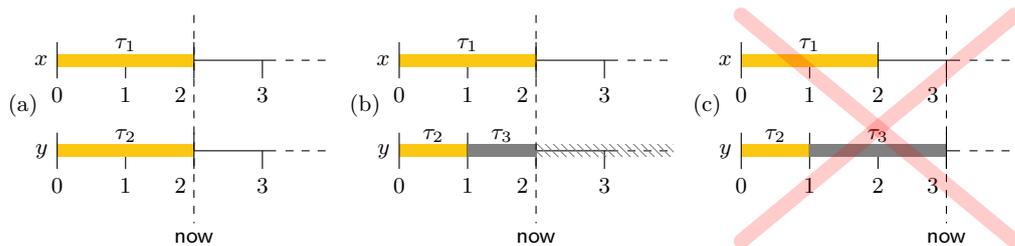

For a partial plan $\Pi=(\Gamma,\now)$, we write
$\T\in\Pi$ to mean $\T\in\Gamma$, and we define $\H(\Pi)=\max_{\T\in\Pi}\H(\T)$
and $\now(\Pi)=\now$. The set of all possible partial plans for a given game $G$
is denoted as $\bfPi_G$, or simply $\mathbf{\Pi}$ where $G$ is understood.

Definition~\ref{def:partialplan} implies that time cannot advance after the end
of a token without specifying its successor, and ensures that the partial plan
has been built only up to $\now$. \Cref{fig:partial-plans} shows an example of
invalid partial plan where a token continues after the end of another and the
successor of the latter is not specified, forming an invalid gap in the
description of the timelines. It also implies that if any empty, thus closed,
timeline is present, then all timelines are empty. Thus, a unique well-defined
\emph{empty} partial plan exists, which we will denote by $\Pi_\emptyset$, with
all empty timelines and $\now(\Pi_\emptyset)=0$.

Players incrementally build a partial plan by extending the initially empty
partial plan $\Pi_\emptyset$. This is done by means of actions that specify
which tokens to start and/or end.

\begin{definition}[Action]
  \label{def:action}
  Let $G=(\SV_C,\SV_E,\S,\D)$ be a planning game and let $SV = SV_C \ \cup \
  SV_E$. An \emph{action} $\alpha$ for $G$ is a term of the form $\tokstart(x,
  v)$ or $\tokend(x,v)$, where $x\in\SV$ and $v\in V_x$. The set of actions is
  partitioned into the set $\mathsf{A}_C$ of \charlie's actions and the set
  $\mathsf{A}_E$ of \eve's actions. An action belongs to $\mathsf{A}_C$ (resp.,
  $\mathsf{A}_E$) if it is of the form $\tokstart(x,v)$, for some $x\in \SV_C$
  (resp., $x\in\SV_E$) and $v\in V_x$, or of the form $\tokend(x,v)$, for some
  $x\in\SV$, $v\in V_x$, and $\gamma_x(v)=\mathsf{c}$
  (resp., $\gamma_x(v)=\mathsf{u}$).\fitpar
\end{definition}

Definition \ref{def:action} can be read as follows. When an action
$\tokstart(x,v)$ or $\tokend(x,v)$ is executed by a player, a token for the
variable $x$, where $x=v$, respectively starts or ends. Ending a task (a token)
and starting the next one are two different actions, even if, as it will be
precisely stated later, the end of a token must be immediately followed by the
start of the next one. Depending on who owns the variable and the involved
value, each action can be executed only by a specific player.
More precisely, players can start tokens for the variables that they own, and
end the tokens that hold values that they control.

It is worth noticing that, in contrast to the original definition of
timeline-based planning problems with uncertainty
(Definition~\ref{def:problem:uncertainty}), Definition \ref{def:action} admits
cases where $x\in\SV_E$ and $\gamma_x(v)=c$ for some $v\in V_x$, that is, cases
where \charlie may control the duration of a variable that belongs to \eve. This
situation is symmetrical to the more common one where \eve controls the duration
of a variable that belongs to \charlie, and there is no need, in our setting, to
impose any asymmetry.

Actions are combined into \emph{moves} that can start and/or end multiple tokens
at once.

\hyphenation{time-stamp}
\begin{definition}[Move]
  \label{def:moves}
  Let $G=(\SV_C,\SV_E,\S,\D)$ be a timeline-based planning game. A~\emph{move}
  $\mu$ is a term of the form $\wait(t)$ or $\play(t, A)$, where $t\in\N$ is the
  \emph{timestamp} of the move, $A \subseteq \A_C$ or $A \subseteq \A_E$, and
  for each $x\in\SV_C\cup\SV_E$, \emph{at most one} action in $A$ involves $x$.
  The set of moves is partitioned into the set $\M_C$ of \charlie's moves and
  the set $\M_E$  of \eve's moves. A move $\mu$ belongs to $\M_C$ if it is
  either of the form $\mu=\wait(t)$ or of the form $\mu=\play(t,A)$ and $A
  \subseteq \A_C$, while it belongs to $\M_E$ only if it is of the form
  $\mu_E=\play(t,A)$ and $A\subseteq\A_E$.\fitpar
\end{definition}

According to Definition \ref{def:moves}, \charlie can either execute some set of
actions $A$, by playing a $\play(t, A)$ move, or wait until some given time $t$,
by playing $\wait(t)$.  In contrast, \eve has a unique kind of move available,
\ie, $\play(t,A)$, which executes the actions in $A$ at time $t$.

We are now ready to introduce the fundamental notion of round.
\begin{definition} [Round]
  \label{def:round}
  A \emph{round} $\rho$ is a pair of moves $(\mu_C,\mu_E)\in\M_C\times\M_E$.
  Let $\Pi$ be a partial plan. A round $\rho=(\mu_C,\mu_E)$ is
  \emph{applicable} to $\Pi$ if the following conditions are met.
  \begin{enumerate}
  \item \label{def:round:applicable:integrity}
        Integrity conditions:
        \begin{enumerate}
        \item any action of the form $\tokstart(x,v)$ is executed by either
              $\mu_C$ or $\mu_E$ if and only if $\T_x\in\Pi$ is a closed
              timeline;
        \item for any action of the form $\tokend(x,v)$, executed by
              $\mu_C$ or $\mu_E$, either an action $\tokstart(x,v)$ is being
              executed by either moves, or $\T_x\in\Pi$ is an open timeline
              $\T_x=\seq{\tau_1,\ldots,\tau_k}$ and its open token is
              $\tau_k=(x,v)$.
        \end{enumerate}
  \item \label{def:round:applicable:timing}
        Timing conditions:
        \begin{enumerate}
        \item if $\mu_C=\play(t_C,A_C)$ and $\mu_E=\play(t_E,A_E)$, then $t_C=t_E=\now(\Pi)$;
        \item if $\mu_C=\wait(t_C)$ and $\mu_E=\play(t_E,A_E)$, then $t_C > \now(\Pi)$ and $t_E\le t_C$.
        \end{enumerate}
  \end{enumerate}
  A single move from either player is \emph{applicable} to $\Pi$ if there exists at least one move from the other player such that the round combining the two moves is applicable to $\Pi$.
\end{definition}

The conditions that make a round applicable can be interpreted as follows.
\cref{def:round:applicable:integrity} ensures that the actions played by each
move of the round are consistent with the current state of the timelines in the
partial plan. In particular, a $\tokstart$ action has to follow a $\tokend$ one,
and $\tokstart$ actions cannot be played on timelines that are already open.
Note that both $\tokstart(x,v)$ and $\tokend(x,v)$ for the same $x$ and $v$ can
be played at the same round, possibly by two different players, provided that
the timeline was previously closed. In this case, the move builds a token of
unitary length (see Definition \ref{def:round:outcome}).
\cref{def:round:applicable:timing} constrains the timings of the moves of a
round: if \charlie does not wait, then he has to play immediately, that is,
$t_C=\now(\Pi)$; otherwise, he can wait until an arbitrary future time point
$t_C$. In both cases, \eve must play at a timestamp $t_E\le t_C$. This
restriction has the following meaning. The advancement of time during the game
is determined mostly by \charlie, who can make it advance one step at the time,
by playing at each round, or skip some time steps at once without playing
anything, by waiting. In both cases, \eve's moves must specify what the
environment is doing, if anything, in the meantime, hence the requirement that
$t_E\le t_C$. If $t_E < t_C$, then time advances only up to $t_E$, so that at
the next round \charlie can timely reply to \eve's move.

The next definition specifies the effects of players' moves.
\begin{definition}[Outcome of a round]
  \label{def:round:outcome}
  Let $\Pi$ be a partial plan and $\rho=(\mu_C,\mu_E)$ be an applicable round.
  The \emph{outcome} of the application of $\rho$ to $\Pi$ is a partial plan
  $\rho(\Pi)$, which is obtained from $\Pi$ by applying the following ordered
  sequence of steps:
  \begin{enumerate}
    \item \label{def:round:outcome:starts}
          for each $\mu\in\{\mu_C,\mu_E\}$, if $\mu=\play(t,A)$, then, for any
          $\tokstart(x,v)\in A$, an open token $\tau=(x,v)$ is appended to
          $\T_x$;
    \item \label{def:round:outcome:ends}
          for each $\mu\in\{\mu_C,\mu_E\}$, if $\mu=\play(t,A)$, then, for any
          $\tokend(x,v)\in A$, the last open token $\tau=(x,v)$ of $\T_x$
          (possibly added at the previous step) is replaced by a closed token
          $\tau'=(x,v,d)$, where $d=t-\starttime(\tau) + 1$;
    \item \label{def:round:outcome:timing}
          $\now(\rho(\Pi))=\min(t_C,t_E)+1$, where either $\mu_C =
          \play(t_C,A_C)$ or $\mu_C = \wait(t_C)$, and $\mu_E=\play(t_E,A_E)$.
  \end{enumerate}
\end{definition}

The effects of $\tokstart$ and $\tokend$ actions are defined in
\cref{def:round:outcome:starts,def:round:outcome:ends}, respectively. The steps
are intended to be applied one after the other in order to handle the case where
both $\tokstart(x,v)$ and $\tokend(x,v)$ are played in the same round. Time
advances according to \cref{def:round:outcome:timing}, depending on the
timestamps of the moves. Note that if no $\wait$ move is played, the new current
time corresponds to the horizon of the resulting partial plan.\fitpar

A play of a planning game is just a sequence of rounds applied to the empty plan
$\Pi_\emptyset$.

\begin{definition}[Play]
  Let $G$ be a planning game and $\Pi_0$ be a partial plan (we call it an
  initial partial plan). A \emph{play} for $G$ from $\Pi_0$ is a sequence
  $\bar\rho=\seq{\rho_0,\ldots,\rho_k}$ of rounds such that $\rho_0$ is
  applicable to $\Pi_0$ and $\rho_{i+1}$ is applicable to
  $\Pi_{i+1}=\rho_i(\Pi_i)$, for $1 \le i < k$.
\end{definition}

Let $\bar\rho=\seq{\rho_0,\ldots,\rho_k}$ be a play and $\Pi$ be a partial plan.
We denote by $\bar\rho(\Pi)=\rho_k(\ldots\rho_0(\Pi))$ the \emph{outcome} of
$\bar\rho$ applied to $\Pi$. Where the initial partial plan is not mentioned, it
is understood that the play is applied to $\Pi_\emptyset$. Each non-empty
partial plan $\Pi$ can be \emph{closed} to form a scheduled plan $\Pi'$ (see
Definition~\ref{def:plan}) by closing all the open tokens of open timelines at
time $\now(\Pi)$. In the following, when the context is clear, we will
interchangeably speak of a partial plan as a scheduled plan by implicitly
referring to its closure. It can be easily checked that rounds as specified in
Definition~\ref{def:moves} suffice to build any possible scheduled plan over the
game  variables, \ie for any partial plan $\Pi$, there exists a play $\bar\rho$
such that $\Pi=\bar\rho(\Pi_\emptyset)$.

\begin{definition}[Strategy for \charlie]
  \label{def:strategy:charlie}
  A \emph{strategy for Charlie} is a function $\sigma_C:\bfPi\to\M_C$ that maps
  any given partial plan $\Pi$ to a move $\mu_C$ applicable to $\Pi$.
\end{definition}

\begin{definition}[Strategy for \eve]
  \label{def:strategy:eve}
  A \emph{strategy for Eve} is a function $\sigma_E:\bfPi\times\M_C\to\M_E$
  that, given a partial plan $\Pi$ and \charlie's move $\mu_C$ applicable to
  $\Pi$, returns the next \eve's move $\mu_E$ such that $\rho=(\mu_C,\mu_E)$
  is applicable to $\Pi$.
\end{definition}

A play $\bar\rho$ is said to be \emph{played according to} a strategy $\sigma_C$
for \charlie, starting from some initial partial plan $\Pi_0$, if
$\rho_i=(\sigma_C(\Pi_{i-1}), \mu_E^i)$, for some $\mu_E^i$, for all
$0<i<|\bar\rho|$, and to be played according to a strategy $\sigma_E$ for \eve
if $\rho_i=(\mu_C^i, \sigma_E(\Pi_{i-1},\mu_C^i))$, for all $0<i<|\bar\rho|$.
For any pair of strategies $(\sigma_C,\sigma_E)$ and any $k\ge0$, there is a
unique run $\bar\rho_k(\sigma_C,\sigma_E)$  of length $k$ played according to
$\sigma_C$ and $\sigma_E$.

It is worth to note that, according to our definition of strategy, \charlie can
base his decisions only on the previous rounds of the game, not including \eve's
move at the current round. Together with the fact that time strictly increases
of at least one time unit at each round, this implies that \charlie has to wait
at least one time unit to react to a move by \eve. This models the realistic
assumption that the \emph{sense-reason-react} loop of the controller needs a
finite amount of time to be executed, and is coherent with the semantics of
dynamic control strategies of flexible plans from \cite{CialdeaMayerOU16}.

Let $G=(\SV_C,\SV_E,\S,\D)$ be a planning game. We define two timeline-based
planning problems, $P_\D=(\SV,\D)$ and $P_G=(\SV,\D\cup\S)$, for $G$. A partial
plan is \emph{admissible} if it is a solution plan for $P_\D$, and
\emph{successful} if it is a solution plan for $P_G$. Similarly a play
$\bar\rho$ is admissible or successful if its outcome $\bar\rho(\Pi_\emptyset)$
is, respectively, admissible or successful.

\begin{definition}[Admissible strategy for \eve]
  A strategy $\sigma_E$ for \eve is \emph{admissible} if for each strategy
  $\sigma_C$ for \charlie, there is a $k\ge 0$ such that the play
  $\bar\rho_k(\sigma_C,\sigma_E)$ is admissible.
\end{definition}

\begin{definition}[Winning strategy for \charlie]
  \label{def:strategy:winning}
  Let $\sigma_C$ be a strategy for \charlie. We say that $\sigma_C$ is a
  \emph{winning strategy} for \charlie if for any \emph{admissible} strategy
  $\sigma_E$ for \eve, there exists an $n\ge0$ such that the play
  $\bar\rho_n(\sigma_C,\sigma_E)$ is successful.
\end{definition}

We say that \charlie \emph{wins} the game $G$ if he has a winning strategy,
while \eve \emph{wins} the game if such a strategy does not exist. Let us
consider again the problem described in \cref{sec:issues} to show a simple
winning strategy. The problem can be viewed as a game with the rules in
\cref{ex:rules} belonging to $\S$ and $\D$ empty. After playing
$\tokstart(x,v_1)$ at the beginning, \charlie only has to wait for \eve to play
$\tokend(x,v_1)$, and then play $\tokstart(x,v_2)$ or $\tokstart(x,v_3)$
according to the current timestamp.

A more involved example can be obtained by considering two variables $x\in\SV_C$
and $y\in\SV_E$, with $V_x=V_y=\{go,stop\}$, unit duration, and rules as
follows.

\begin{center}
  \begin{array}{r@{} r @{} r @{${} \implies {}$} l @{} l}
    \multirow{2}{*}[-1pt]{$\S =$}
      & \ldelim\{{2}{*}
      & a[x=stop]
      & \exists b[y=stop] \suchdot b \ebefores_{[0,0]} a & \rdelim\}{2}{*} \\
      & & \true     & \exists a[x=stop] \suchdot \true     &    \\[0.5em]
    \D = & \bigl\{ & \true & \exists a[y=stop] \suchdot \true   & \bigr\}
  \end{array}
\end{center}

Here, \charlie's ultimate goal is to realise $x=stop$, but this can only happen
after \eve realised $y=stop$. This is guaranteed to happen, since we consider
only admissible strategies. Hence, the winning strategy for \charlie only
chooses $x=go$ until \eve chooses $y=stop$, and then wins by executing $x=stop$.
If $\D$ was instead empty, a winning strategy would not exist since a strategy
that never chooses $y=stop$ would be admissible. This would therefore be a case
where \charlie looses because \eve can indefinitely postpone his victory.

Let us compare now the concept of \emph{dynamic controllability} of flexible
plans, as defined in~\cite{CialdeaMayerOU16}, with the existence of winning
strategies for timeline-based planning games, and show the greater generality of
the latter concept.
\ifpreprint\else Proofs are omitted because of space concerns, but are
available in an extended report~\cite{GiganteMCOR18}.\fi

The first step is to back the claim of this greater generality. Indeed, it can
be shown that for any timeline-based planning problem with uncertainty
$P=(\SV_C,\SV_E,S,\O)$ there is an \emph{associated timeline-based planning
game} $G_P=(\SV_C,\SV_E,\D,\S)$, such that a dynamically controllable flexible
solution plan for $P$ gives us a winning strategy for $\G$. To this aim, we need
a way to represent as a game any given planning problem with uncertainty.
Intuitively, this can be done by encoding the observations $\O$ into suitable
domain rules. The game associated with a problem therefore mimics the exact
setting described by the problem. What follows shows that such a game does
indeed exist, and that there is a close relationship between its winning
strategies and the dynamically controllable flexible plans for the problem.

\begin{restatable}{theorem}{planstostrategiesthm}
  \label{thm:planstostrategies}
  Let $P$ be a timeline-based planning problem with uncertainty. If $P$
  admits a \emph{dynamically controllable flexible solution plan}, then \charlie
  has a winning strategy for the associated timeline-based planning game $G_P$.
\end{restatable}

\Cref{thm:planstostrategies} shows that the proposed framework can represent any
timeline-based planning problem with uncertainty, and that the notion of winning
strategy for a game subsumes that of dynamically controllable flexible plan.
Moreover, it can be seen that the game setting is strictly more expressive, \ie
there are cases where no dynamically controllable flexible plans exist, but a
winning strategy can be found. This is the case with the example problem
discussed in \cref{sec:issues}, which has an easy winning strategy when seen as
a game, while it has no dynamically controllable flexible plan. Therefore, one
can prove the following theorem.\fitpar

\begin{restatable}{theorem}{noplansthm}
  \label{thm:noplans}
  There exists a timeline-based planning problem with uncertainty
  $P=(\SV_C,\SV_E,S,\O)$ such that there are no dynamically controllable
  flexible plans for $P$, but \charlie has a winning strategy for the associated
  planning game $G_P$.
\end{restatable}

\section{Decidability and complexity}
\label{sec:decidability}

A natural question about  timeline-based planning games is whether a winning
strategy for a given game can effectively be found. In this section, we
positively answer the question, showing that a winning strategy, if it exists,
can be found with an algorithm that runs in doubly exponential time.
Proofs are omitted because of space concerns, but are available
in Gigante's Ph.D.\ thesis~\cite{Gigante19}.

The proposed algorithm makes use of the concept of \emph{concurrent game
structure} (CGS)~\cite{AlurHK02}, in particular, of the specific subclass of
\emph{turn-based synchronous game structures} (simply \emph{game structures},
from now on), that, in the case of two players only, can be defined as follows.

\begin{definition}[Turn-based synchronous game structure~\cite{AlurHK02}]
  A \emph{turn-based synchronous game structure} is a tuple
  $S=(Q_1, Q_2, M_1, M_2,\P,\pi,\delta_C,\delta_E)$,
  where:
  \begin{enumerate}
  \item $Q_1$ and $Q_2$ are the finite sets of \emph{states} belonging to
        Player~1 and Player~2, respectively;
  \item $M_1$ and $M_2$ are the finite sets of \emph{moves} available to
        Player~2 and Player~2, respectively;
  \item $\P$ is a finite set of \emph{proposition letters};
  \item $\pi:Q_1\cup Q_2\to 2^\P$
        provides the set of proposition letters that hold at any given state;
  \item $\delta_1:Q_1\times M_1\times Q_2$ and
        $\delta_2:Q_2\times M_2\times Q_1$ are the \emph{transition relations},
        that specify the states reachable from a given one by applying a move by
        either player.
  \end{enumerate}
\end{definition}

We proceed in two steps. We first provide a suitable encoding of a
timeline-based planning game $G$ into a corresponding game structure, and then
we exploit existing machinery for $\ATL*$ model checking \cite{AlurHK02} to find
a winning strategy for $G$.

Let us focus on the first step. To encode a timeline-based planning game into a
game structure, one may think of interpreting partial plans as states, marking
with some proposition letter $d$ those which satisfy the domain rules and with
$w$ those where the system rules are satisfied.
However, such an encoding can, in principle, lead to an infinite state space, as
the number of possible partial plans is infinite: the set of moves available to
each player is infinite, as they specify a timestamp; moreover, at any given
time, a rule triggered by one of the current tokens may require to look
arbitrarily back in the past in order to check (the possibility of) its
satisfaction. We now show how to constrain the state space to be finite.

First of all, we observe that arbitrarily large timestamps can only be
introduced by $\wait$ moves played by \charlie, as the timestamp of any $\play$
move played by \charlie is forced to be equal to $\now$ and the timestamp of any
\eve's move is bounded by the one of a \charlie's move. Now, even though $\wait$
moves are useful for \charlie to skip a given amount of time without executing a
sequence of empty $\play$ moves, they can always be replaced by such a sequence,
and thus it can be easily shown that if a winning strategy exists, another one
exists which does not use any $\wait$ move. Hence, w.l.o.g., we can safely
restrict ourselves to a \emph{finite} set of moves of form $\play(t,A)$, with
$t=\now$.\fitpar

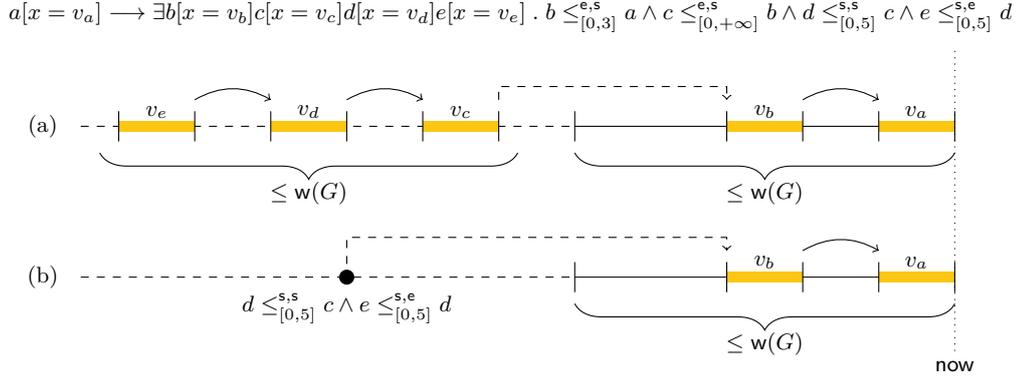
\begin{figure}
  \footnotesize
  \begin{equation*}
    a[x=v_a] \implies \exists b[x=v_b] c[x=v_c] d[x=v_d] e[x=v_e]\suchdot
      b\ebefores_{[0,3]} a \land c \ebefores_{[0,+\infty]} b \land
      d \sbefores_{[0,5]} c \land e \sbeforee_{[0,5]} d
  \end{equation*}

  \centering

  \begin{tikzpicture}[every node/.append style={math mode}]
    \def\h{3cm}
    \def\pad{0.2cm}
    \def\w{12cm}

    \coordinate (tx) at ({2*\pad},{2/3*\h-\pad});
    \coordinate (window) at ($(tx)+({\w-5cm},0)$);
    \coordinate (endx) at ($(tx)+(\w,0)$);

    \node (x) at (tx) {\text{(a)}};
    \draw[dashed] (x) ++(0.5cm,0) -- (window);
    \draw (window) -- (endx);

    \tick at (window);

    \draw[decorate,decoration={brace,mirror,raise=10pt,amplitude=10pt}]
    (window) -- (endx) node[midway,yshift=-25pt] {\le \window(G)};

    \token a at ($(endx)-(1cm,0)$) of color lipics, length 1cm,
      and label {v_a};
    \token b at ($(sa)-(2cm,0)$) of color lipics, length 1cm, and label {v_b};

    \draw[->] (eb) ++(0,10pt) to [bend left] ++(1cm,0);

    \token c at (varx) +(5cm,0) of color lipics, length 1cm, and label {v_c};
    \token d at (varx) +(3cm,0) of color lipics, length 1cm, and label {v_d};
    \token e at (varx) +(1cm,0) of color lipics, length 1cm, and label {v_e};

    \draw[->, dashed] (ec) +(0,10pt) |- ++(0,15pt) -| ($(sb)+(0,10pt)$);
    \draw[->] (ed) +(0,10pt) to [bend left] ($(sc)+(0,10pt)$);
    \draw[->] (ee) +(0,10pt) to [bend left] ($(sd)+(0,10pt)$);

    \draw[decorate,decoration={brace,mirror,raise=10pt,amplitude=10pt}]
      ($(se)+(-0.25cm,0)$) -- ($(ec)+(0.25cm,0)$)
        node[midway, yshift=-25pt] {\le \window(G)};

    \coordinate (cx) at ($(tx)+(0,-2cm)$);
    \coordinate (cwindow) at ($(cx)+({\w-5cm},0)$);
    \coordinate (cendx) at ($(cx)+(\w,0)$);

    \node (xx) at (cx) {\text{(b)}};
    \draw[dashed] (xx) ++(0.5cm,0) -- (cwindow);
    \draw (cwindow) -- (cendx);

    \tick at (cwindow);

    \draw[decorate,decoration={brace,mirror,raise=10pt,amplitude=10pt}]
    (cwindow) -- (cendx) node[midway,yshift=-25pt] {\le \window(G)};

    \token ca at ($(cendx)-(1cm,0)$) of color lipics, length 1cm,
      and label {v_a};
    \token cb at ($(sca)-(2cm,0)$) of color lipics, length 1cm, and label {v_b};

    \draw[->] (ecb) ++(0,10pt) to [bend left] ++(1cm,0);

    \draw (cx) ++(4cm,0)
      node[circle,fill=black,inner sep=2pt,
           label=below:{d \sbefores_{[0,5]} c \land e \sbeforee_{[0,5]} d}]
      (component) {};

    \draw[->, dashed] (component) +(0,5pt) |- ++(0,15pt) -| ($(scb)+(0,10pt)$);

    \draw[dotted] ($(endx)+(0,1cm)$) -- ($(cendx)+(0,-1cm)$) node[below] {\now};
  \end{tikzpicture}

  \caption{%
    How to represent the distant history of a play of a game $G$ in a compact
    way: (a) the complete description of the timeline; and (b) the compact
    description of the distant history.%
  }
  \label{fig:window}
\end{figure}

Let us focus now on the partial plans to show that not every single detail of
their distant past has to be remembered. Consider a timeline-based planning game
$G=(\SV_C,\SV_E,\S,\D)$, and let the \emph{window} of $G$, $\window(G)$, be the
product of all the non-zero lower and upper bounds appearing in any bounded atom
of any rule, and all the non-zero lower and upper bounds on the length of tokens
of any state variable. The window of the game gives a coarse upper bound on how
far from a given point a chain of bounded atoms can look, which is a fundamental
parameter of any timeline-based planning problem, \eg it has been exploited in
\cite{GiganteMCO17} to study the complexity of the plan existence problem.

For the sake of clarification, consider the above half of
Figure~\ref{fig:window}, which shows a timeline satisfying a particular
synchronisation rule. The satisfaction of bounded and unbounded atoms of the
rule by tokens are represented by solid and dashed arrows, respectively. The
depicted tokens can be partitioned in two blocks, which satisfy two different
groups of bounded atoms. Once the first group has been witnessed, only its
existence need to be remembered, not the precise position or length of its
tokens, since any bounded atom involving the trigger cannot be affected by
anything farther than the window $\window(G)$. Hence, while the events happening
at most $\window(G)$ time steps before $\now$ need to be remembered in detail,
the satisfaction of the other group of atoms can be remembered symbolically as
shown in the bottom half of Figure~\ref{fig:window}, \ie by remembering the
existence of the tokens satisfying each required group of atoms, and their
relative ordering. The same has to be done to remember future requests triggered
by tokens appeared in the past.

It can be easily checked that $\window(G)\in\O(2^{\abs{G}})$, and that the
number of bits needed to store the succinct representation of the history and of
future requests is exponential as well.  Hence, we can build a data structure of
size at most \emph{exponential} in the size of $G$, that, during the whole play,
can be used to represent the current state and the past history of the game.

Let $\Pi$ be a partial plan. We denote by $[\Pi]$ its succinct representation,
and by $[\bfPi_G]$ the set of the succinct representations of all the possible
partial plans on $G$. For any round $\rho$, we can easily specify its
application to the succinct representation of a partial plan in such a way that
$\rho([\Pi])=[\rho(\Pi)]$. Similarly, given $[\Pi]$, we can easily check whether
$\Pi$ is a solution plan for a given timeline-based planning problem.

We are now ready to define the \emph{game structure} $\G$ associated with a
planning game $G$.\fitpar

\begin{definition}[Game structure for a planning game]
  \label{def:game-structure-encoding}
  Let $G=(\SV_C,\SV_E,\S,\D)$ be a timeline-based planning game. The corresponding game structure
  $\G$ is a tuple $(Q_C,Q_E,M_C,$ $M_E,\P,\pi,\delta_C,\delta_E)$, where:
  \begin{enumerate}
  \item $Q_C=[\bfPi_G]$ and $Q_E\subseteq[\bfPi_G]\times\M_C$;
  \item $M_C=\M_C$ and $M_E=\M_E$;
  \item $\P=\{d,w\}$, where $d$ and $w$ are proposition letters,
  \item $\pi(q_E)=\emptyset$, for any $q_E\in Q_E$, and $d\in\pi([\Pi])$
  (resp., $w\in\pi([\Pi])$ if and only if $\Pi$ is a solution plan for $P_\D$ (resp., $P_G$);
  \item $([\Pi],\mu_C,([\Pi],\mu_C))\in\delta_C$ if and only if $\mu_C=\play(t,A)$ is
        applicable to $\Pi$ and $t=\now$;
  \item $(([\Pi],\mu_C),\mu_E,[\Pi'])\in\delta_E$ if and only if $\rho=(\mu_C,\mu_E)$ is
        applicable to $\Pi$ and $\rho([\Pi])=[\Pi']$.
  \end{enumerate}
\end{definition}

The winning condition for \charlie (see Definition~\ref{def:strategy:winning})
can be expressed by means of the following \ATL* formula, where Player~1 is
interpreted as \charlie and Player~2 as \eve:%
\footnote{%
  \ATL* is a branching-time temporal logic similar to \CTL*, that allows one to
  quantify over the strategies of the different players of a game structure and
  to differentiate between paths played according to those strategies.%
}
\begin{equation*}
    \phi\equiv \sE{1}\bigl(\F d \to \F(d\land w)\bigr)
\end{equation*}
The formula asks for the existence of a strategy $\sigma_C$ for \charlie
($\sE{1}$) such that, for all paths played according to $\sigma_C$, if the path
is played according to an admissible strategy for \eve ($\F d$), then there is a
future point where $d\land w$ holds ($\F(d\land w)$), \ie \charlie wins. By
applying known \ATL* model-checking algorithms on the game structure encoding
the planning game and the above formula, one may solve the problem.\fitpar

It can be proved that any strategy satisfying the above \ATL* formula
corresponds to a winning strategy for \charlie on the original timeline-based
planning game. The proposed compact representation of partial plans has
exponential size, and thus the game structure has a doubly exponential number of
states. Since \ATL* model-checking of fixed-size formulae over game structures
is \PTIME-complete~\cite{AlurHK02}, the following theorem holds.
\begin{theorem}
  \label{thm:winning-strategies}
  Let $G$ be a timeline-based planning game. The problem of establishing whether
  \charlie has a winning strategy on $G$ belongs to
  \textnormal{\textsf{2}}\EXPTIME.
\end{theorem}

\section{Conclusions and future work}
\label{sec:conclusions}

This paper defines \emph{timeline-based planning games}, a novel game-theoretic
approach to timeline-based planning with uncertainty. Unlike current
formulations based on dynamic controllability of flexible plans, the proposed
one can uniformly deal with both temporal uncertainty and general
nondeterminism, and it is strictly more expressive. We showed that a winning
strategy can be found in doubly exponential time. Whether there exists a
matching complexity lower bound, how to synthesise a finite-state machine
implementing such a strategy, and how hard the synthesis problem is are still
open issues.

\bibliography{biblio}

\end{document}